\documentclass[twoside,journal]{IEEEtran}
\usepackage{amsmath,amsfonts}
\usepackage{algorithmic}
\usepackage{algorithm}
\usepackage{array}
\usepackage[caption=false,font=normalsize,labelfont=sf,textfont=sf]{subfig}
\usepackage{textcomp}
\usepackage{stfloats}
\usepackage{url}
\usepackage{verbatim}
\usepackage{graphicx}
\usepackage{cite}
\usepackage{times}
\usepackage{soul}
\usepackage{url}
\usepackage[hidelinks]{hyperref}
\usepackage[utf8]{inputenc}
\usepackage{graphicx}
\usepackage{amsmath}
\usepackage{amsthm}
\usepackage{booktabs}
\usepackage{algorithm}
\usepackage{algorithmic}
\usepackage{color}
\usepackage{multirow}
\usepackage{array}
\usepackage{amssymb}
\usepackage{booktabs}

\newcommand{\wanglong}[1]{#1}
\newcommand{\revise}[1]{#1}
\newcommand{\proofread}[1]{#1}

\newcommand{\nrevise}[1]{#1}

\begin{document}

\title{GRIG: Few-Shot Generative Residual\\Image Inpainting}

\author{Wanglong~Lu,
	Xianta~Jiang,
	Xiaogang~Jin,~\IEEEmembership{Member,~IEEE,}
	Yong-Liang~Yang,\\
 	Minglun~Gong,
	Tao~Wang,
	Kaijie~Shi,
	and~Hanli~Zhao$^{*}$

\thanks{W. Lu and H. Zhao are with the Key Laboratory of Intelligent Informatics of Safety \& Emergency of Zhejiang Province, Wenzhou University, Wenzhou 325035, China.}
\thanks{W. Lu, X. Jiang, and K. Shi are with the Department of Computer Science, Memorial University of Newfoundland, St. John's, NL A1B 3X5, Canada.}
\thanks{X. Jin is with the State Key Laboratory of CAD\&CG, Zhejiang University, Hangzhou 310058, China.}
\thanks{Y.-L. Yang is with the Department of Computer Science, University of Bath, Bath, BA2 7AY, United Kingdom.}
\thanks{M. Gong is with the School of Computer Science University of Guelph Guelph, ON, N1G 2W1, Canada.}
\thanks{T. Wang is with the Department of Computer Science and Technology, Nanjing University, China.}
\thanks{$^{*}$ Corresponding author. E-mail: hanlizhao@wzu.edu.cn}

}



\maketitle

\begin{abstract}
Image inpainting is the task of filling in missing or masked region of an image with semantically meaningful contents. Recent methods have shown significant improvement in dealing with large-scale missing regions. However, these methods usually require large training datasets to achieve satisfactory results and there has been limited research into training these models on a small number of samples. To address this, we present a novel few-shot generative residual image inpainting method that produces high-quality inpainting results. The core idea is to propose an iterative residual reasoning method that incorporates Convolutional Neural Networks (CNNs) for feature extraction and Transformers for global reasoning within generative adversarial networks, along with image-level and patch-level discriminators. We also propose a novel forgery-patch adversarial training strategy to create faithful textures and detailed appearances. Extensive evaluations show that our method outperforms previous methods on the few-shot image inpainting task, both quantitatively and qualitatively.
\end{abstract}


\begin{figure*}[t]
	\centering
	\includegraphics[width=\textwidth]{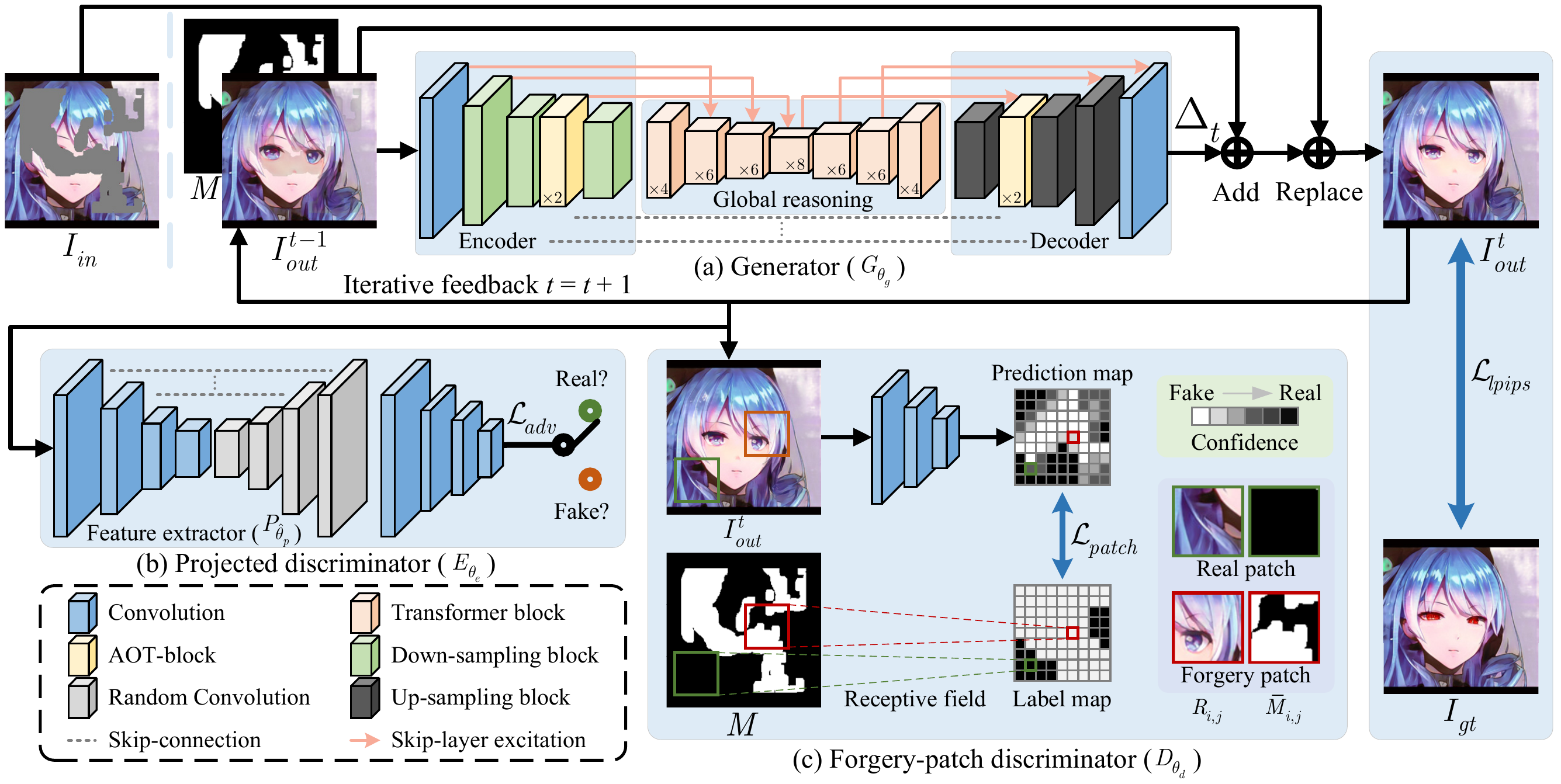}
	\caption{Overall pipeline of our few-shot generative residual image inpainting framework (GRIG). For each $t$-{th} residual reasoning step, the generator \proofread{(a)} utilizes the $(t-1)$-{th} inpainted image $I_{out}^{t-1}$ to generate the residual image $\Delta_{t}$. At the first residual reasoning step ($t=1$), we set $I_{out}^0 = I_{in}$. With the inpainted image $I_{out}^{t-1}$ and the initial input image $I_{in}$, the add and replace operations are performed to obtain the $I_{out}^{t}$ for the next iterative refinement. During adversarial training, the inpainted image is fed into the projected discriminator \proofread{(b)} and forgery-patch discriminator \proofread{(c)}, respectively. At each iterative reasoning step, the loss functions and corresponding back-propagation are re-computed. During the test phase, a similar multi-step prediction is performed without the loss functions and back-propagation. For the sake of simplicity, down- and up-sampling operations are omitted.
 }\label{fig:fig_residual_inpainting}
\end{figure*}

\section{Introduction}
\label{sec:intro}
\IEEEPARstart{I}{mage} 
 inpainting is a fundamental task in computer graphics~\cite{Mori2020} and computer vision~\cite{Li2022}. It has been employed in many downstream applications, such as diminished reality~\cite{Kawai2016,Mori2022}, image restoration~\cite{Wan2020}, and image manipulation~\cite{Jo2019}.
Recently proposed image inpainting methods have achieved impressive results~\cite{Suvorov2022,Li2022} on both realistic and facial images. However, these methods have an overlooked limitation: they require a large amount of data to train the Convolutional Neural Network (CNN) or Transformer models~\cite{Vaswani2017}. When these models are trained on small image datasets, there is a high possibility of overfitting and model collapse~\cite{Arjovsky2017_ICLR}. \revise{In practice, it is much easier for users if only a small number of training images are required. On the other hand,} large image datasets in certain domains (e.g., medical, art, and historical relics images) are either too expensive or infeasible to collect. This has severely restricted the use of image inpainting in real-world scenarios. \revise{Furthermore, improving the data efficiency of model training can be a crucial factor in expediting its adoption and application, thereby increasing accessibility for various fields that face limitations in data availability.} 

Achieving high-quality image inpainting on few-shot datasets is still a challenging and open problem. \revise{Most of existing methods~\cite{Zhao2021,Li2022,Yu2019} rely on single-pass inference which may generate ambiguous results in inpainted sub-regions. Some methods~\cite{Zeng2020,Li2020} perform inpainting in a progressive fashion by reusing parts of previously inpainted features from early refinement stages. However, these methods do not fully utilize the inpainted pixels as useful information for the next iteration. Moreover, existing inpainting methods are not designed \proofread{specifically} for few-shot learning and may not work well with limited training samples. Domain-related prior knowledge~\cite{Ojha2021} or lightweight generative models~\cite{Liu2020,Sauer2021} may be employed in image inpainting to mitigate overfitting. However, they do not perform well when there is a large domain gap between two tasks or the reduced network capacity affects the inpainting quality.}

\wanglong{To this end,} we propose a novel few-shot Generative Residual Image inpaintinG framework (GRIG), which enables high-quality image inpainting on few-shot datasets. \wanglong{To effectively optimize inpainting results, we use iterative reasoning to solve more accurate and generalizable algorithmic reasoning tasks~\cite{du2022irem} \proofread{along with} residual learning~\cite{He2016} to incrementally refine previous estimates.
We \proofread{investigate} whether combining iterative reasoning and residual learning with CNNs and Transformers~\cite{Vaswani2017},} \revise{as well as decoupled image-level and patch-level discriminators,} \wanglong{can lead to a promising design for creating a more robust and data-efficient method to tackle the few-shot image inpainting task.} In this framework, the inpainting process is carried out in several forward passes by feeding the generator with the output of the previous iteration and a corresponding mask. This enables the generator to be well-trained and focused on the residual information between the previously predicted output and the ground truth, resulting in effective image inpainting with high quality and good generality on few-shot data.


\revise{We implemented our framework into three components: a generator, a projected discriminator, and a forgery-patch discriminator.}  The generator uses CNNs to extract shallow features of edges and textures and the Transformer blocks to capture global interactions between feature contexts at each iterative step. To accelerate network convergence and reduce overfitting, \revise{we decouple the image distribution learning by using image-level and patch-level discriminators. We first build the projected discriminator to capture the whole image-level distribution}. We then propose a forgery-patch discriminator for enhancing the patch-level details of generated images, as the projected discriminator has difficulty in capturing fine details in inpainted images.  Experimental results on ten few-shot datasets show that our method is superior to the state-of-the-art (SOTA) methods in terms of few-shot and high-quality image inpainting.

In summary,  this paper makes the following contributions:
\begin{itemize}
    \item A novel few-shot generative residual image inpainting framework that integrates CNNs and Transformers, and decoupled image-level and patch-level discriminators for enhanced performance on few-shot data.
 
	\item A forgery-patch discriminator that assists the generative network to improve the fine details of generated images and prevent overfitting for few-shot image inpainting.
	
	\item State-of-the-art performance on ten few-shot benchmark datasets with varying contents and characteristics, including facial, photorealistic, animal, medical, cartoon, and art-like images.
\end{itemize}

\section{Related work}\label{relatedwork}
{Early image inpainting techniques} 
rely heavily on low-level features from pixels and image patches. Methods based on diffusion~\cite{Ballester2001,Bertalmio2000,Levin2003} propagate undamaged information along the boundary to the hole's center. Patch-based methods~\cite{Telea2004,Kwatra2005,Guo2018,Zhu2016_tvcg} iteratively search for and copy similar appearances from image datasets or known backgrounds. Some variants include GPU-based parallel methods~\cite{Barnes2009}, summarizing of non-stationary patterns~\cite{Simakov2008}, and inpainting with nonlocal texture similarity~\cite{Ding2019}. 
Because of the lack of semantic understanding of the image, these methods perform well for small-scale and narrow missing regions but fail to recover meaningful contents for large holes.

{Deep-learning-based inpainting methods} have achieved great success in semantic completion. Deep neural networks have been used extensively to improve the visual quality of inpainting~\cite{Ren2015}. These works include an auto-encoder-based architecture~\cite{Pathak2016} and its variant architectures~\cite{Wang2018,Yan2018,Zeng2019}. Various sophisticated modules or learning strategies have been developed to enhance the effectiveness of image inpainting, including global and local discriminators~\cite{Iizuka2017}, contextual attention~\cite{Liu2019,Xie2019,Yi2020,Yu2018} to improve semantic understanding, methods for dealing with irregular holes~\cite{Liu2018,Yu2019,Zhao2021}, and utilization of auxiliary information (such as sketches~\cite{Jo2019}, foreground contours~\cite{Xiong2019}, and structures~\cite{Ren2019}).  Recent research has addressed issues related to high-resolution~\cite{Yi2020,Zeng2021,Suvorov2022,Li2022,Zheng2021,Zeng2020}, 3D photography~\cite{Mori2020}, pluralistic generation~\cite{Li2022,Zhao2021,Zhao2020,Wan2021,Liu2022}, and large hole filling~\cite{Zhang2018_inpaint,Li2019,Li2020,Li2022,Zhao2021,Wan2021,Zheng2021,Liu2022,Guo2019_inpaint}. The methods discussed above aim for semantically high-quality completion, but they may overfit when trained on data with  a small number of samples.

Progressive-based image inpainting methods~\cite{Li2019,Zeng2020,Li2020,Zhang2018_inpaint,Guo2019_inpaint} are closely related to our work. These methods primarily inpaint pixels from the hole boundary to the center in a progressive manner~\cite{Li2019,Zhang2018_inpaint,Guo2019_inpaint,Zeng2020} or employ multi-stage refinement schemes~\cite{Zeng2020,Li2020}. For example, Zeng et al.~\cite{Zeng2020} improved high-resolution inpainting by iteratively predicting a confidence map and corresponding intermediate results. Such methods reuse only a portion of the predicted information and do not change pixels with high confidence for the next iterative inpainting. Recurrent Feature Reasoning (RFR)~\cite{Li2020} runs embedded feature maps through their feature reasoning module multiple times to generate multiple features for adaptive feature merging. RFR's final inpainted results, on the other hand, are produced from their decoder with a single forward pass, indicating that the model cannot readjust its results at the pixel level for better fine details. In this paper, we make a first attempt at image inpainting training on a small number of samples. We show how our framework can refine the results of inpainting \wanglong{through} iterative residual reasoning, \wanglong{which ingeniously combines CNNs and Transformers,} \revise{as well as image-level and patch-level discriminators.}  \wanglong{Our approach  allows for the efficient reuse of} all previously predicted pixels and network parameters, \wanglong{leading to an effective model for the few-shot image inpainting task}.

\section{Methodology}\label{meth}

\wanglong{Iterative reasoning, which involves applying underlying computations to the outputs of previous reasoning steps repeatedly, has the potential to solve more accurate and generalizable algorithmic reasoning tasks~\cite{du2022irem}, whereas residual learning~\cite{He2016,Hur2019} facilitates the progressive optimization of previously predicted results. By iteratively predicting residual offsets and reusing previously predicted information within the same generator during each reasoning step, our GRIG} \revise{can dynamically learn to refine the input image at each step. This method avoids memorizing the mapping between the inputs and their ground-truth images, thereby preventing overfitting and improving visual quality for few-shot inpainting.}

As shown in Fig.~\ref{fig:fig_residual_inpainting}, our framework consists of three main parts: a generator, a projected discriminator, and a forgery-patch discriminator. Given a ground-truth image $I_{gt} \in \mathbb{R}^{h \times w \times 3}$ and a binary mask $M \in \mathbb{R}^{h \times w \times 1}$ (with 1 for unknown and 0 for known pixels), the masked image $I_{in} \in \mathbb{R}^{h \times w \times 3}$ is obtained as $I_{in} = I_{gt} \odot (1-M)$, where $\odot$ denotes the Hadamard product. The goal of GRIG is to automatically inpaint a realistic image $I_{out}^{T} \in \mathbb{R}^{h \times w \times 3}$ with $T$ steps of iterative reasoning, where $T>1$ denotes the iterative reasoning  steps during training. For each $t$-th iterative residual reasoning step, a previously inpainted image $I_{out}^{t-1}$ is fed into the generator to obtain a residual prediction. At the first residual reasoning step ($t=1$), we set $I_{out}^0 = I_{in}$. Then, addition and replacement operations are performed to produce a new image completion $I_{out}^t$. The adversarial training is conducted at each iterative step with the network weights updated accordingly via back-propagation.

\subsection{Network architectures}

\subsubsection{Generator} 
\wanglong{Taking the previous iteration's inpainted results as input, the generator is designed to combine CNNs and Transformers~\cite{Vaswani2017} for efficient iterative residual reasoning} \revise{in few-shot image inpainting.} The generator $G_{\theta_{g}}$ consists of an encoder, a \revise{global reasoning module with a} stack of Restormer's Transformer blocks~\cite{Zamir2022}, and a decoder (see Fig.~\ref{fig:fig_residual_inpainting}a). \wanglong{The CNN-based encoder and decoder excel at feature extraction, whereas the Transformer blocks excel at dynamic attention, global context integration, and generalization.} \revise{This combination helps the generator generalize effectively on few-shot training samples.} To extract features and enlarge the receptive field for capturing both informative distant image contexts and rich patterns of interest, we first stack a convolution layer, several residual down-sampling blocks~\cite{Liu2020}, and AOT-blocks~\cite{Zeng2021}. The extracted features will then be fed into a Restormer's Transformer block stack for global context reasoning. Meanwhile, skip-layer excitation modules (SLE)~\cite{Liu2020} are utilized for a shortcut gradient flow, and skip connections are employed for collecting the multi-resolution feature maps in the decoder. 
The decoder is then built using up-sampling blocks~\cite{Liu2020}, AOT-blocks~\cite{Zeng2021}, and a convolution layer. The decoder generates the intermediate prediction $\Delta_{t}$ by utilizing the multi-resolution feature maps output by the encoder and \revise{global reasoning module}. For stable adversarial training, we apply spectral normalization~\cite{Miyato2018} to all convolution layers of the networks.

At each $t$-th iterative reasoning step, the inpainted image from the previous iteration $I_{out}^{t-1}$ and its corresponding mask $M$ (see Fig.~\ref{fig:fig_residual_inpainting}a) are fed into a generative network $G_{\theta_{g}}$ with the learnable network parameters ${\theta}_{g}$. $G_{\theta_{g}}$  generates the intermediate residual inpainting $\Delta_{t} = G_{\theta_{g}}(I_{out}^{t-1}, M) \in \mathbb{R}^{h \times w \times 3}$. Then, the $t$-{th} inpainted image $I_{out}^{t}$ is calculated as:
\begin{equation}\label{equ:g_forward}
	\begin{aligned}
		I_{out}^{t} = (I_{out}^{t-1}+\Delta_{t}) \odot M + I_{in} \odot (1-M).
	\end{aligned}
\end{equation}

\begin{figure*}[t]
  \centering
  \includegraphics[width=0.8\textwidth]{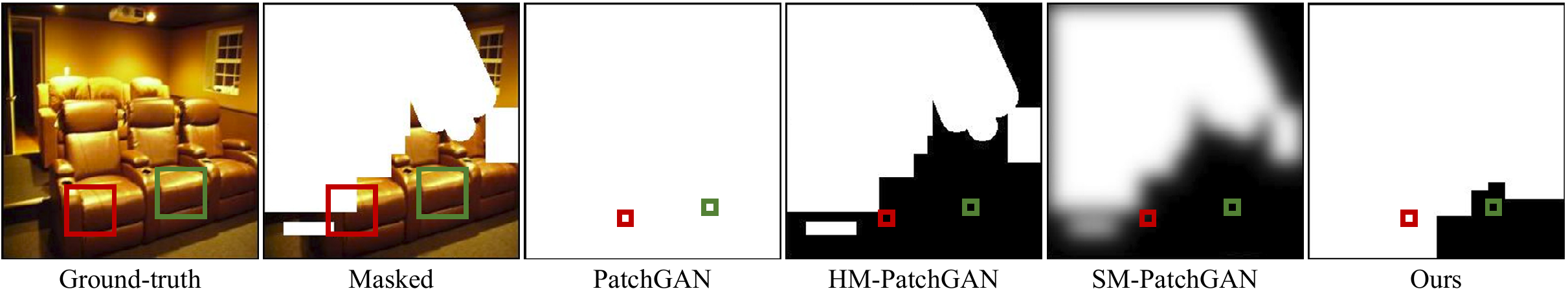}
  \caption{\wanglong{Illustration of the differences between the discriminators of PatchGAN~\cite{Isola_2017_cvpr}, HM-PatchGAN~\cite{Zeng2021}, SM-PatchGAN~\cite{Zeng2021}, and our algorithm. The boxes represent patches with the size of the discriminator's receptive field (left two images) and  corresponding projected positions (right four images) in the resultant label maps over the (red) masked and (green) unmasked regions; Pixel values in the label maps indicate labels for fake (white) and real (black) patches.}}
  \label{fig:label_map_comparison}
\end{figure*}

\subsubsection{Projected discriminator} 
\wanglong{To stabilize GAN training and improve data efficiency, we use prior knowledge from pre-trained representations to train a compact classifier \proofread{for learning} the global distribution of few-shot images.} The projected discriminator (see Fig.~\ref{fig:fig_residual_inpainting}b) learns to assign high confidence scores to feature maps extracted from real images while assigning low scores to synthesized ones.
Initially, feature maps are extracted from the input image $I$ (i.e., $I_{out}^{t}$ or $I_{gt}$) using a U-net-like projector $P_{\hat{\theta}_{p}}$ with the pre-trained network parameters $\hat{\theta}_{p}$. $P_{\hat{\theta}_{p}}$ is implemented by a pre-trained EfficientNet-Lite1~\cite{Tan2019} with cross-channel mixing and cross-scale mixing mechanisms~\cite{Sauer2021}. Subsequently, the projected discriminator $E_{\theta_{e}}$ with the learnable network parameters $\theta_{e}$ maps the extracted feature maps to a scalar. Here we selected the discriminator with the largest scale of feature projections (i.e., removing the other three small-scale discriminators) from Projected GAN~\cite{Sauer2021}.

\subsubsection{Forgery-patch discriminator} Because the projected discriminator is primarily focused on extracting global image features for robust classification, it is possible that some fine detail features may be overlooked in these projected features.  To help the generator produce faithful fine-grained textures \revise{and avoid overfitting in few-shot training}, we propose a forgery-patch discriminator that learns to identify real and inpainted patches based on the receptive field~\cite{Yu2016} of the discriminator.

As shown in Fig.~\ref{fig:fig_residual_inpainting}c, the forgery-patch discriminative network $D_{\theta_{d}}$ with the learnable network parameters $\theta_{d}$ learns to recognize real or forgery image patches from a given image $I$ (i.e., $I_{out}^{t}$ or $I_{gt}$). The discriminator $D_{\theta_{d}}$ maps $I$ to a prediction map, where each unit indicates a confidence score for each image patch based on the receptive field. In this work, we adopted the network architecture of $D_{\theta_{d}}$ from PatchGAN~\cite{Isola_2017_cvpr}.  \revise{The patch-level receptive field in neural networks has also been studied \proofread{as} a means of overfitting avoidance in interactive video stylization~\cite{Texler2020SIG} and improving diversity and generalizability in image generation~\cite{wang2022sindiffusion}.}

\subsection{Objective functions}
GRIG is trained to optimize the learnable network parameters $\theta_{g}$, $\theta_{e}$, and $\theta_{d}$ using the objective functions listed below. 
\subsubsection{LPIPS loss}
At each iterative reasoning step, we use the Learned Perceptual Image Patch Similarity (LPIPS) metric~\cite{Zhang2018_lpips} to constrain the perceptual similarity between the inpainted image $I_{out}^{t}$ and the ground-truth image $I_{gt}$:
\begin{equation}\label{equ:loss_lpips}
	\begin{aligned}
		\mathcal{L}_{lpips}(\theta_g)=  \| F(I_{out}^{t})-F(I_{gt})\|_{2},
	\end{aligned}
\end{equation}
where $F(\cdot)$ is the pre-trained perceptual feature extractor, and we use VGG-16 in our work~\cite{Simonyan2014}. This can assist our generative network in learning to maintain higher visual quality.

\subsubsection{Projected adversarial loss}
For fast convergence, the projected adversarial loss utilizes pre-trained classification models to extract prior knowledge (see Fig.~\ref{fig:fig_residual_inpainting}b).
 We employ the hinge loss~\cite{Sauer2021} to optimize the projected discriminator $E_{\theta_{e}}$ and generative network $G_{\theta_g}$, respectively. The objective function can be formulated as:
\begin{equation}\label{equ:loss_adv}
	\begin{aligned}
		\mathcal{L}_{adv}^{E}(\theta_e) &= 
		\mathbb{E}_{I_{gt}}[ReLU(1-E_{\theta_{e}}(P_{\hat{\theta}_{p}}(I_{gt})))] \\ &+\mathbb{E}_{I_{out}^{t}}[ReLU(1+E_{\theta_{e}}(P_{\hat{\theta}_{p}}(I_{out}^{t})))], \\ 
		\mathcal{L}_{adv}^{G}(\theta_g) &= -\mathbb{E}_{I_{out}^{t}}[E_{\theta_{e}}(P_{\hat{\theta}_{p}}(I_{out}^{t}))]. \\
	\end{aligned}
\end{equation}

The projected discriminator is constrained to assign low  scores to inpainted images and high scores to real images, while the generator $G_{\theta_g}$ is supervised by the projected discriminator to inpaint the masked input based on the distribution of real images.

\subsubsection{Adversarial forgery-patch loss}

As shown in Fig.~\ref{fig:fig_residual_inpainting}c, we enforce the forgery-patch discriminator to  distinguish forgery and real patches in a given image. We achieve this by constructing the corresponding label map $X \in \mathbb{R}^{h' \times w'} $ to supervise the discriminator. Specifically, we partition $I$ and $M$ into $h' \times w'$ pairs of partially overlapped patches ($R_{i,j}$ and $\overline{M}_{i,j}$) based on the receptive field of forgery-patch discriminator $D_{\theta_{d}}$. 
Here, $1 \leq i \leq h'$ and $1 \leq j \leq w'$ are horizontal and vertical indices, and the sizes of $R_{i,j}$ and $\overline{M}_{i,j}$ are equal to the receptive field $N \times N$.  The label map is expressed as follows:
\begin{equation}\label{equ:label_map}
	\begin{aligned}
		X_{i,j} = \begin{cases} 0 & \text {if } \| \overline{M}_{i,j} \|_{0} = 0;
			\\1 & \text {otherwise,} \end{cases}
	\end{aligned}
\end{equation}
where $\| \overline{M}_{i,j} \|_{0}$ is defined as the L0 norm of the sub-region mask $\overline{M}_{i,j}$. If the $\| \overline{M}_{i,j} \|_{0}$ is not zero, it shows there are some masked pixels in this sub-region mask, and the image patch $R_{i,j}$ contains inpainted pixels. Thus, we set $X_{i,j} = 1$, which means that the sub-region $R_{i,j}$ is a forgery patch. Otherwise, it is a real patch. The hinge version of adversarial forgery-patch loss is expressed as:
\begin{equation}\label{equ:transition_3}
	\resizebox{.91\linewidth}{!}{$
		\begin{aligned}
		\mathcal{L}_{patch}^{D}(\theta_{d}) &=\mathbb{E}_{{I}_{gt}}[ReLU(1-D_{\theta_{d}}(I_{gt}))] \\
		&+\mathbb{E}_{I_{out}^t}[ReLU(1-D_{\theta_d}(I_{out}^{t}))\odot (1-X)]\\
		&+\mathbb{E}_{I_{out}^t}[ReLU(1+D_{\theta_d}(I_{out}^{t}))\odot X], \\
		\mathcal{L}^{G}_{patch}(\theta_g) &= -\mathbb{E}_{I_{out}^{t}}[D_{\theta_{d}}(I_{out}^{t})\odot X].
	\end{aligned}
	$}
\end{equation}

\wanglong{Fig.~\ref{fig:label_map_comparison} illustrates the differences between the proposed forgery-patch discriminator and closely related discriminators. }
PatchGAN's discirminator~\cite{Isola_2017_cvpr} directly assigns all patches in inpainted images as fake patches, which can confuse the discriminator when extracted patches do not have any generated pixel.  \wanglong{HM-PatchGAN} and SM-PatchGAN~\cite{Zeng2021} aim to segment synthesized patches of missing regions according to inpainting masks. Since the inpainting masks have to be downsampled first to align with the spatial size of the discriminator's output, the constraints around the mask boundaries may be unclear. For example, downsampling on inpainting masks results in information loss on the precise location of inpainted pixels.  SM-PatchGAN tries to identify the generated and real patches, whereas our discriminator goes one step further to consider whether generated pixels are consistent with surrounding real pixels in a given patch.  \wanglong{Our discriminator constructs the label map based on the receptive field and} treats all patches with any inpainted pixels as fake patches, which gives more constraints than PatchGAN and SM-PatchGAN.

\begin{figure*}[t]
	\centering
	\includegraphics[width=1\textwidth]{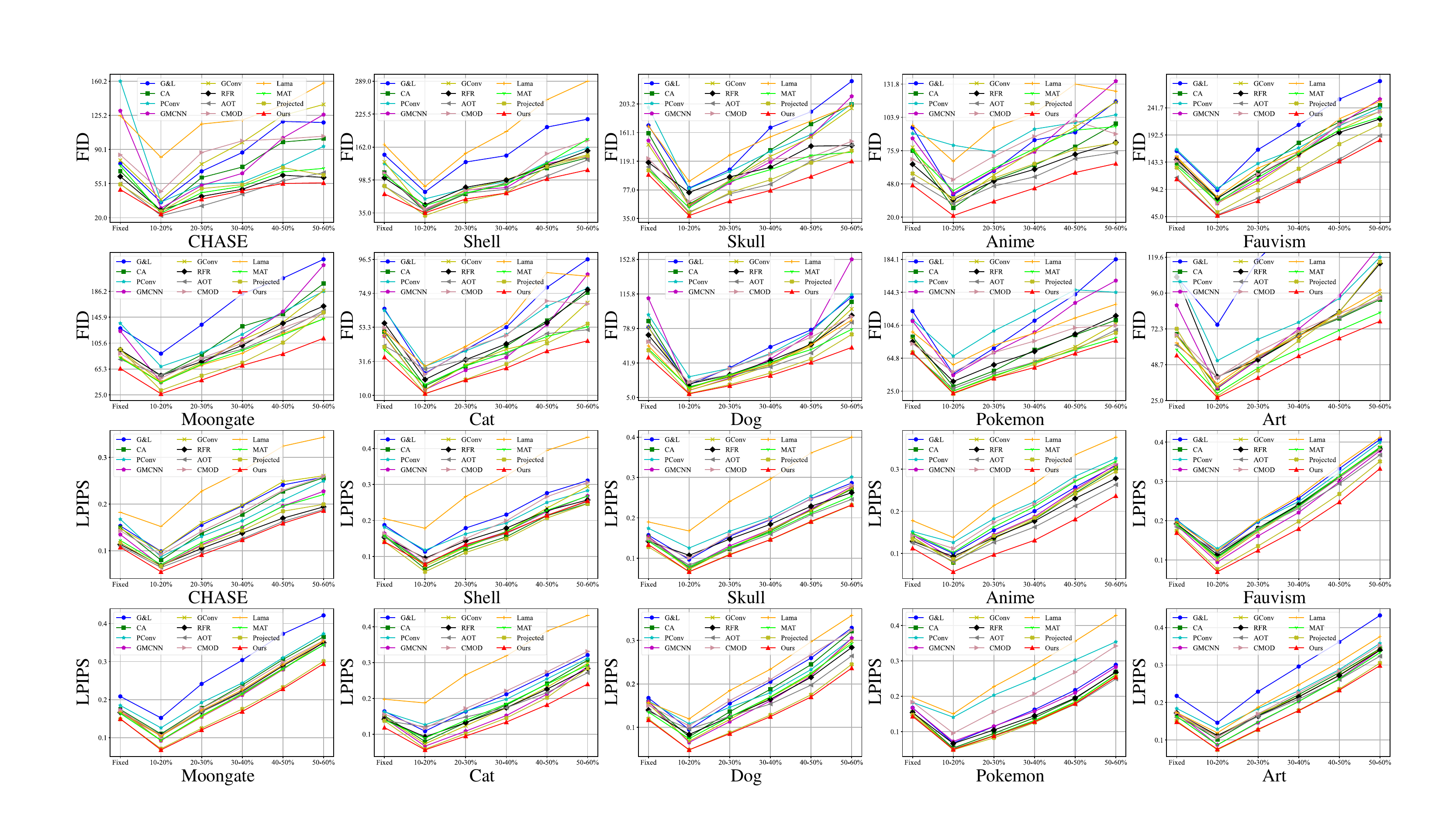}
	\caption{Quantitative comparisons of GRIG with the SOTA image inpainting methods on ten few-shot datasets and two evaluation metrics (FID in top two rows and LPIPS in bottom two rows). ``Fixed'' denotes the fixed center 25\% rectangular mask.}\label{fig:fig_few_shot_quantative_comparision_fid_lpips}
\end{figure*}

\subsubsection{Total objective}\label{sec:total_losses}
The total training objective of the generator is expressed as:
\begin{equation}\label{equ:loss_total}
	\begin{aligned}
		\mathcal{L}_{total}^{G} = \lambda_{lpips}\mathcal{L}_{lpips} + \lambda_{adv}\mathcal{L}_{adv}^{G} +  \lambda_{patch}\mathcal{L}_{patch}^{G},
	\end{aligned}
\end{equation}
where $\lambda_{lpips}$,  $\lambda_{adv}$, and $\lambda_{patch}$ are weights of corresponding losses, respectively. During training, we alternately optimize parameters ${\theta}_{g}$, ${\theta}_{e}$, and ${\theta}_{d}$.

\begin{algorithm}[t]
	\caption{Training procedure of GRIG}
	\label{alg:procedure_framework}
	\begin{algorithmic}[1] 
		\WHILE{$G_{\theta_g}$, $E_{\theta_e}$, and $D_{\theta_d}$ have not converged}
		\STATE Sample batch images $\mathcal{I}_{gt}$ from the training set
		\STATE Create random masks $\mathcal{M}$ for $\mathcal{I}_{in}$
		\STATE Get inputs $\mathcal{I}_{in} \leftarrow \mathcal{I}_{gt} \odot (1 - \mathcal{M}) $
		\STATE Set inputs $\mathcal{I}_{out}^{0} \leftarrow \mathcal{I}_{in} $
		\FOR {iterative residual reasoning step $t=1$ to $T$}
		\STATE Get $\Delta_{t} \leftarrow  G_{\theta_g}\left(\mathcal{I}_{out}^{t-1}, \mathcal{M} \right)$
		\STATE Get $\mathcal{I}_{out}^{t} \leftarrow (I_{out}^{t-1}+ \Delta_{t})  \odot \mathcal{M}+ \mathcal{I}_{in} \odot (1 - \mathcal{M})$ 
		\STATE Update $G_{\theta_g}$ with $\mathcal{L}_{total}^{G}$
		\STATE Update $E_{\theta_e}$ with  $\mathcal{L}_{adv}^{E}$
		\STATE Update $D_{\theta_d}$ with  $\mathcal{L}_{patch}^{D}$
		\ENDFOR
		\ENDWHILE
	\end{algorithmic}
\end{algorithm}

\subsection{Iterative residual  reasoning}

The iterative residual  reasoning for image inpainting can be formulated as an optimization process over adversarial generative networks. This enables the generator to implicitly learn to leverage previously predicted results and focus on residual information in order to achieve high-quality and better generality.

We introduce a generative network  $G_{\theta_g}(I_{out}^{t-1})$ as an explicit function to predict residual information (see Eq.~\ref{equ:g_forward}). At each iterative reasoning step $t$, the generator $G_{\theta_g}$ is trained to maximize the confidence values of $E_{\theta_e}(I_{out}^t)$ and values in the prediction map of $D_{\theta_d}(I_{out}^t)$ while minimizing the perceptual similarity between $I_{out}^t$ and $I_{gt}$.  Thus, $\theta_g$ is solved by $\theta_g = {\arg \min }_{{\theta_g}} \mathcal{L}_{total}^{G}$. Simultaneously,  $E_{\theta_e}$ and $D_{\theta_d}$ are trained to distinguish images (fake or real) and patches, respectively, where  ${\theta}_{e}=\arg \min _{\theta_{e}} \mathcal{L}_{adv}^{E}$ and ${\theta}_{d}=\arg \min _{\theta_{d}} \mathcal{L}_{patch}^{D}$. The generative network $G_{\theta_g}$ directly predicts the residual information while its parameters $\theta_g$ are supervised by the discriminators as well as $I_{gt}$. \wanglong{The pseudo-code for the GRIG training procedure is listed in Algorithm~\ref{alg:procedure_framework}.}

\begin{figure*}[!thbp]
	\centering
	\includegraphics[width=1\textwidth]{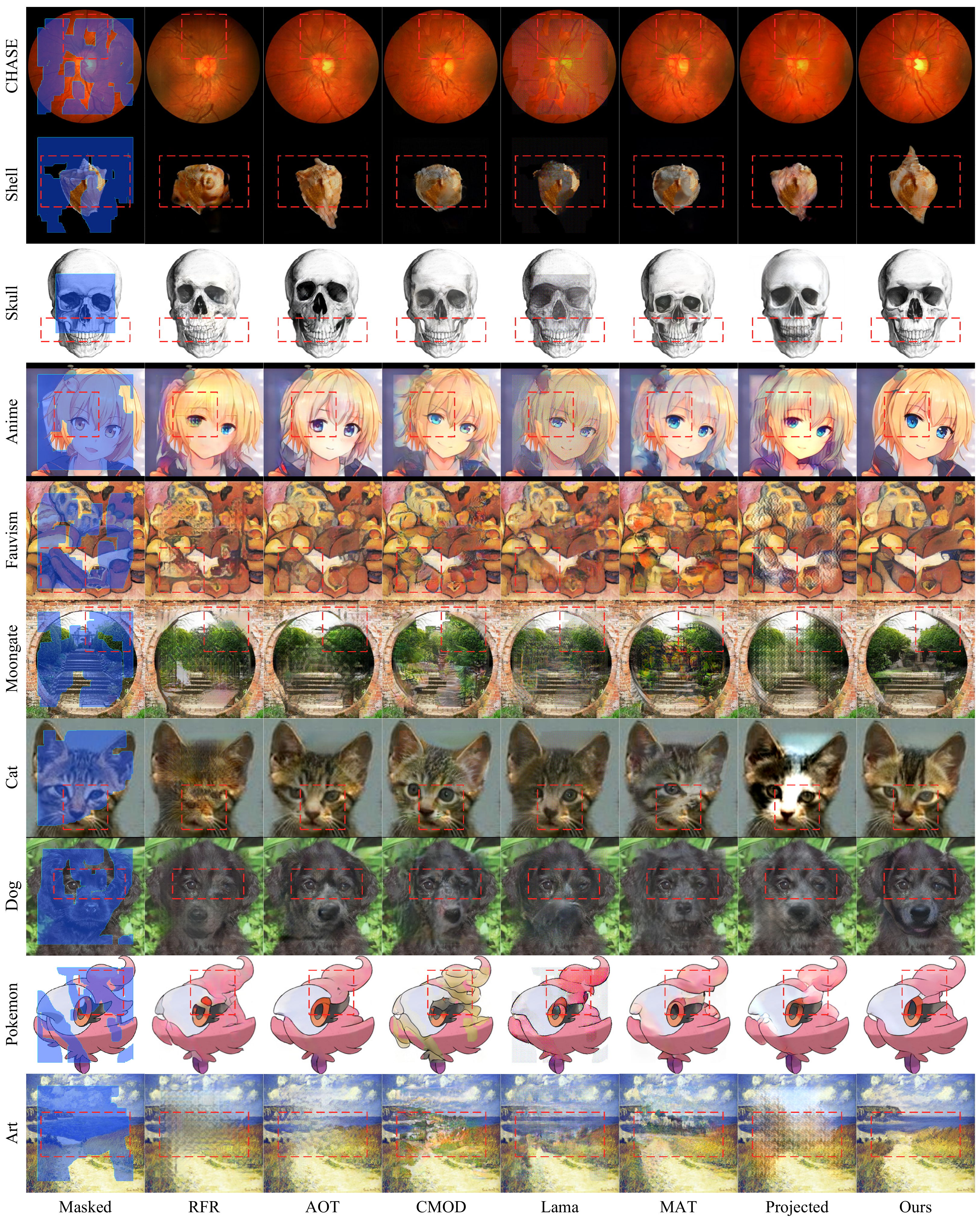}
	\caption{Visual comparisons of GRIG and SOTA image inpainting methods on few-shot datasets.
}\label{fig:fig_inpainting_performance_sub_b}
\end{figure*}

\begin{figure*}[!htbp]
	\centering
	\includegraphics[width=0.8\textwidth]{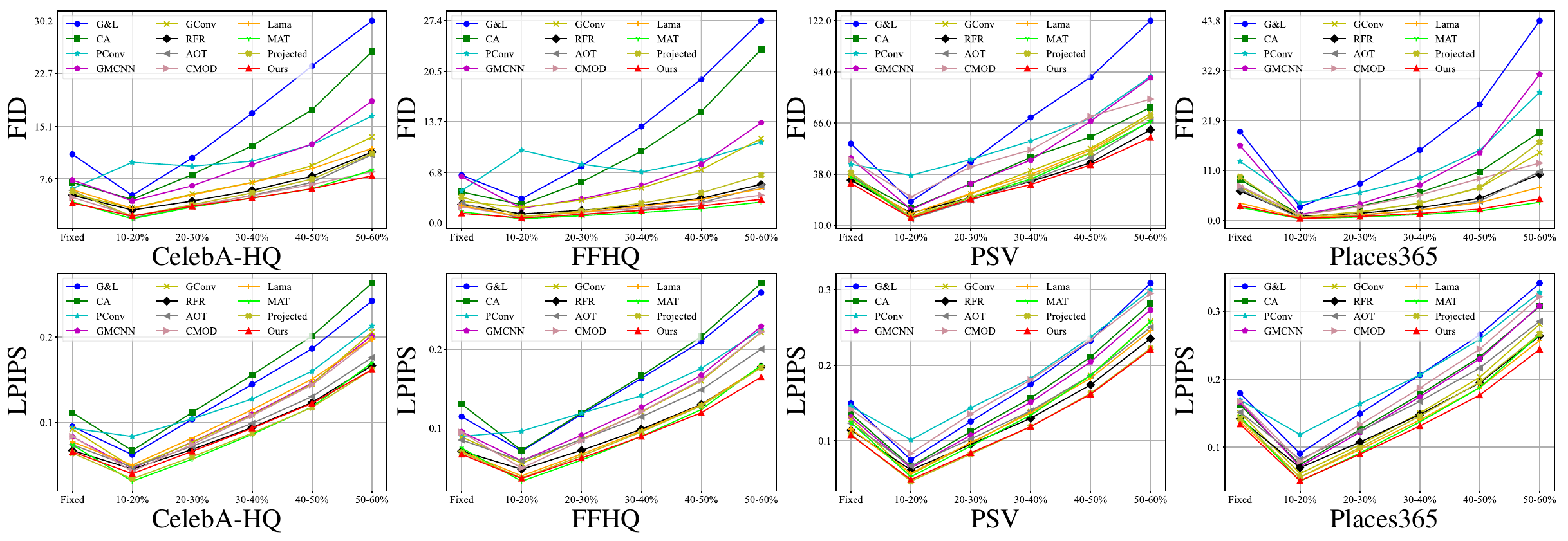}
	\caption{Quantitative comparisons of GRIG with the SOTA image inpainting methods on four large-scale datasets and two evaluation metrics
(FID and LPIPS). “Fixed” denotes the fixed center 25\% rectangular mask.}\label{fig:fig_large_quantative_comparision}
\end{figure*}

\section{Experiments}
\label{exp}

\subsection{Experimental setup}\label{sec:implementation}

Python and PyTorch were used to build the proposed framework. \wanglong{We set $\lambda_{lpips}= 1.5$, $\lambda_{adv}= 1$, $\lambda_{patch} = 1$, and $T=3$ for all experiments in both training and testing phases, unless otherwise specified.} We used the Adam optimizer with first momentum coefficient $\beta_1=0.5$, second momentum coefficient $\beta_2=0.999$, and learning rate 0.0002. Our masks were created with the CMOD mask generation algorithm~\cite{Zhao2021}. \nrevise{Our generator contains 31.76M parameters and achieves around $21$ FPS for each residual reasoning step on the NVIDIA GeForce RTX 2080 GPU (8 GB).}

We compared GRIG to SOTA image inpainting methods, including Globally\&Locally (G\&L)~\cite{Iizuka2017}, Contextual Attention (CA)~\cite{Yu2018}, Partial Convolutions (PConv)~\cite{Liu2018}, GMCNN~\cite{Wang2018}, Gated Convolution (GConv)~\cite{Yu2019}, Recurrent Feature Reasoning (RFR)~\cite{Li2020}, AOT-GAN (AOT)~\cite{Zeng2021}, Co-mod-GAN (CMOD)~\cite{Zhao2021}, Lama~\cite{Suvorov2022}, and MAT~\cite{Li2022}. \revise{We also compared GRIG to an inpainting model (Projected) based on the light-weight Projected GAN~\cite{Sauer2021} to further demonstrate the superiority of GRIG on few-shot inpainting.} The publicly available MMEditing framework~\cite{mmediting2022}, which is an open-source image and video editing toolbox based on PyTorch, \revise{implements the models of G\&L~\cite{Iizuka2017}, CA~\cite{Yu2018}, PConv~\cite{Liu2018}, and GConv~\cite{Yu2019}. We used the official codes of GMCNN~\cite{Wang2018}, RFR~\cite{Li2020}, AOT~\cite{Zeng2021}, Lama~\cite{Suvorov2022}, and MAT~\cite{Li2022}. We used the official TensorFlow-based code to create a PyTorch-based version of CMOD~\cite{Zhao2021}. To implement the Projected model, we added a mirrored encoder of Projected GAN~\cite{Sauer2021} with skip connections and the perceptual similarity $\mathcal{L}_{lpips}$. This Projected model was created using PyTorch with the same hyper-parameters of GRIG with $\lambda_{lpips}= 1.5$.}

To ensure fairness, we used the same training/testing splits for all experiments. All images were resized to the resolution of $256 \times 256$. All compared models were retrained on the datasets mentioned in the paper, \nrevise{using a batch size of 8}, unless otherwise noted. All models were trained \nrevise{and tested} on NVIDIA V100 GPUs (32 GB).
\nrevise{ During testing, various irregular masks with different mask ratios~\cite{Liu2018} and a fixed center 25\% ($128 \times 128$) rectangular mask were used to simulate different situations for all experiments. }

Since L1 distance, PSNR, and SSIM all heavily prefer blurry results~\cite{Zhao2021}, we used Fr\'{e}chet inception distance (FID)~\cite{Heusel2017} and LPIPS metrics for quantitative evaluation following the established practice in the recent literature~\cite{Suvorov2022}.

\begin{table}[t]
	\centering
    \setlength{\tabcolsep}{5mm}
    \caption{The detailed information of the ten experimented few-shot image datasets.}
 {
		\begin{tabular}{c|ccccc}
			\hline
			Dataset & \# Training set & \# Test set\\
            \hline
			CHASE~\cite{Fraz2012}   &  18   &    10    \\ 
            \hline
            shell~\cite{Liu2020}    & 48    & 16\\
            \hline
			skull~\cite{Liu2020}   &  72   &    25    \\
            \hline
            Anime~\cite{Liu2020}    & 90    & 30\\
            \hline
			Fauvism~\cite{Liu2020}   &  94   &    30    \\ 
            \hline
            moongate~\cite{Liu2020}    & 106    & 30\\
            \hline
			Cat~\cite{Zhu_Si2011}   &  120   &    40    \\ 
            \hline
            Dog~\cite{Zhu_Si2011}    & 309    & 80\\
            \hline
			Pokemon (pokemon.com)   &  633   &    200    \\ 
            \hline
            Art (wikiart.org)    & 750    & 250\\
			\hline
	\end{tabular}}
	\label{tab:datasets}
\end{table}

\subsection{Comparison results on few-shot datasets}\label{sec:small_comparison}

Experiments were conducted on ten few-shot image datasets. The detailed information of the datasets is as shown in Table \ref{tab:datasets}. All datasets were split by random sampling. \nrevise{All models were trained with $400,000$ image batches.}

\revise{Fig.~\ref{fig:fig_few_shot_quantative_comparision_fid_lpips} shows quantitative comparisons of GRIG with the SOTA image inpainting methods on ten few-shot datasets.}
To compare the performance of image inpainting, various irregular masks with different mask ratios~\cite{Liu2018} as well as a fixed center 25\% ($128 \times 128$) rectangular mask were used to simulate various scenarios. 
In the few-shot setting, the differentiable data-augmentation~\cite{Zhao2020_NIPS} was applied for all compared methods when sampling images in the training phase.  \revise{As shown in Fig.~\ref{fig:fig_few_shot_quantative_comparision_fid_lpips},} GRIG outperforms all baselines in terms of FID and LPIPS metrics by large margins on most benchmarks for various kinds of masks. For most datasets, significant gains were obtained by our method. 
Notably, under 50-60\% mask ratio,  GRIG achieves a relative improvement of FID to the second-best methods of $10.03\%$ (CHASE), $16.26\%$ (Shell), $11.03\%$ (Skull), $14.49\%$ (Anime), $4.24\%$ (Fauvism), $26.48\%$ (Moongate), $15.61\%$ (Cat), $23.86\%$ (Dog), $3.70\%$ (Pokemon), and $7.31\%$ (Art). 

Fig.~\ref{fig:fig_inpainting_performance_sub_b} presents \wanglong{the inpainted results of the compared methods. It reveals that most methods fail to produce plausible contents for datasets with fewer than 100 training samples (for example, CHASE, shell, skull, and anime) due to overfitting to features and patterns from a small number of samples. When trained on datasets with more than 500 samples (such as Pokemon and Art), some methods may be able to fill more semantic content within masked areas. However, artifacts can still be seen under close inspection.} When the masked area is large, RFR~\cite{Li2020} is prone to producing repetitive image patches in inpainted regions. While AOT~\cite{Zeng2021} and Lama~\cite{Suvorov2022} can inpaint structures in missing regions, they leave artifacts in fine details. CMOD~\cite{Zhao2021} and MAT~\cite{Li2022}  tend to overfit the training data due to their large amounts of learnable parameters.  Projected GANs~\cite{Sauer2021} can handle the semantic structure, but they may introduce color inconsistency around mask boundaries. Fig.~\ref{fig:fig_few_shot_quantative_comparision_fid_lpips} and Fig.~\ref{fig:fig_inpainting_performance_sub_b} demonstrate that GRIG can achieve better performance on quantitative metrics and visual quality, even though our method has more learnable parameters (31.76M) than those of GConv~\cite{Yu2019} (4.0M) and is trained on limited samples. \revise{GRIG demonstrates strong generalization capabilities in various few-shot datasets with differing training sample numbers and produces images with higher visual quality.}

\begin{figure*}[!htbp]
	\centering
	\includegraphics[width=1\textwidth]{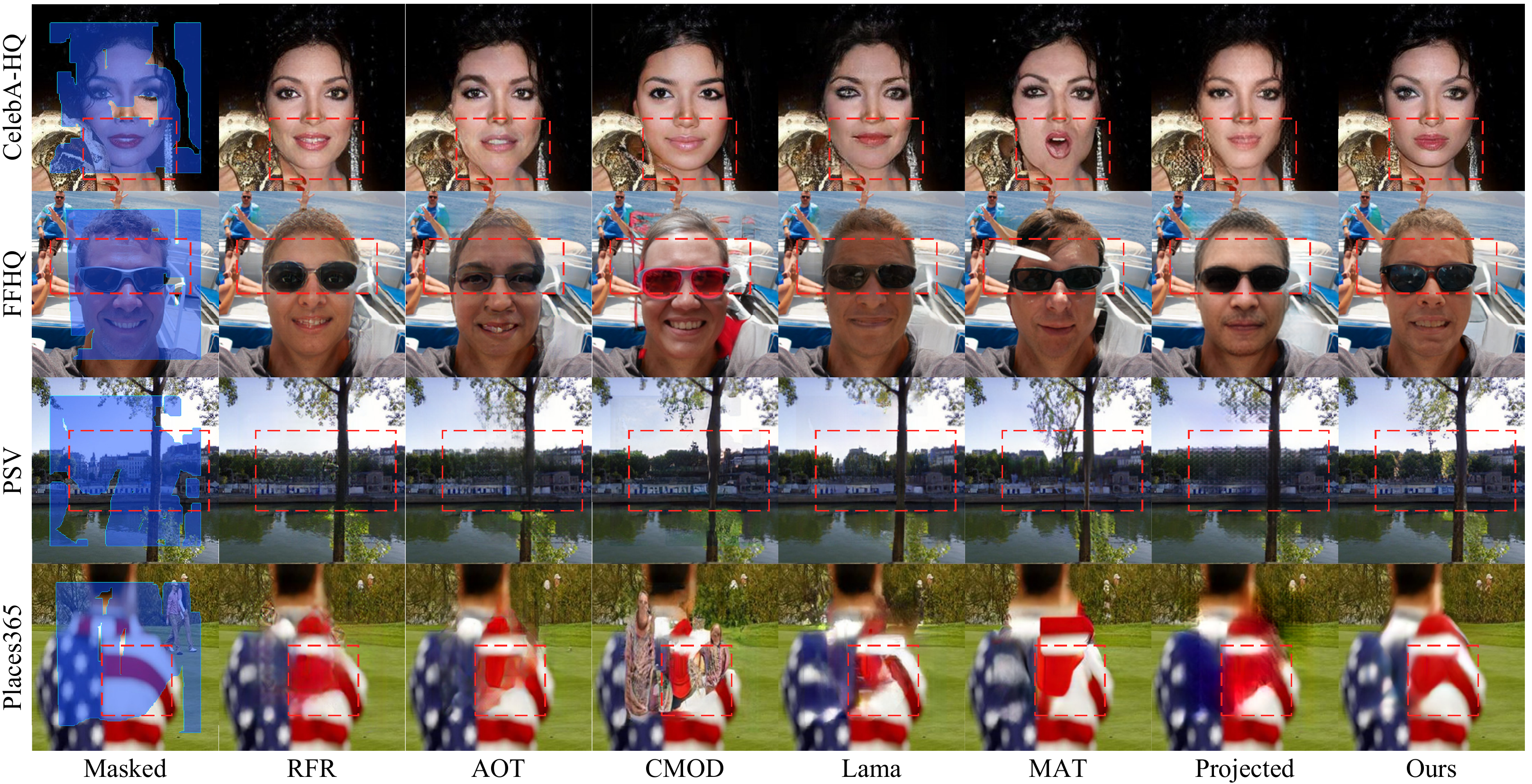}
	\caption{Visual comparisons of the GRIG and SOTA image inpainting methods on large-scale datasets.}\label{fig:fig_large_free_form_img}
\end{figure*}

We believe that there are three reasons for the better generalization  performance and inpainting quality achieved \revise{on few-shot image inpainting}. First, our iterative residual reasoning strategy enables the generator to use information learned in previous iterations while also increasing the diversity of inputs to improve performance. Second, the self-attention mechanism in Transformers~\cite{Vaswani2017} has advantages in leveraging existing information for further context reasoning. In our generator, the encoder and decoder are used to extract local features, while the Restormer Transformer blocks~\cite{Zamir2022} are used for global context reasoning. Third, the projected discriminator and forgery-patch discriminator, with 2.829M  and 2.765M learnable parameters, respectively, help improve the generality of our method. The projected discriminator focuses on images at the semantic level based on the generality of pre-trained features. The forgery-patch discriminator \revise{focuses on learning patch-level consistency to capture patch statistics and distinguishing between real and inpainted patches to prevent overfitting by avoiding the need to memorize the entire image}.

\subsection{Comparison results on large-scale datasets}

We also compared our method to SOTA inpainting methods on four large-scale datasets: CelebA-HQ~\cite{Karras2018} (28K for training, 2K for testing), FFHQ~\cite{Karras2019} (60K for training, 10K for testing), Paris Street View~\cite{Doersch2012} (PSV, 14.9K for training, 100 for testing), Places365~\cite{Zhou2017} (1.8M for training, 36.5K for testing), respectively. We used the original training and testing splits from the PSV and Places365 datasets directly, while other datasets were split using random sampling. All methods were trained with their default settings to ensure fair comparisons. \nrevise{Our model was trained with $1,000,000$ image batches on CelebA-HQ, FFHQ, and PSV, respectively, and $2,000,000$ image batches on Places365.}

The quantitative results in Fig.~\ref{fig:fig_large_quantative_comparision} show that GRIG outperforms the majority of SOTA inpainting methods in terms of FID and LPIPS metrics on large-scale datasets. In particular, GRIG achieves the best FID scores on PSV, and the best LPIPS scores on PSV and Places365. MAT~\cite{Li2022} has the best FID scores on FFHQ and Places365. Overall, GRIG performs comparably to MAT on the other large-scale datasets while containing much fewer learnable weights (31.76M) than MAT (62.0M). \revise{Our iterative residual learning effectively assists the networks in decomposing the inpainting process into multiple reasoning steps with the progressive refinement of inpainting results. Moreover, the decoupling of image distribution learning into image-level and patch-level constraints with our projected discriminator and forgery-patch discriminator helps our GRIG model achieve excellent performance in both few-shot scenarios and large datasets.}

\begin{figure}[t]
	\centering
	\includegraphics[width=0.485\textwidth]{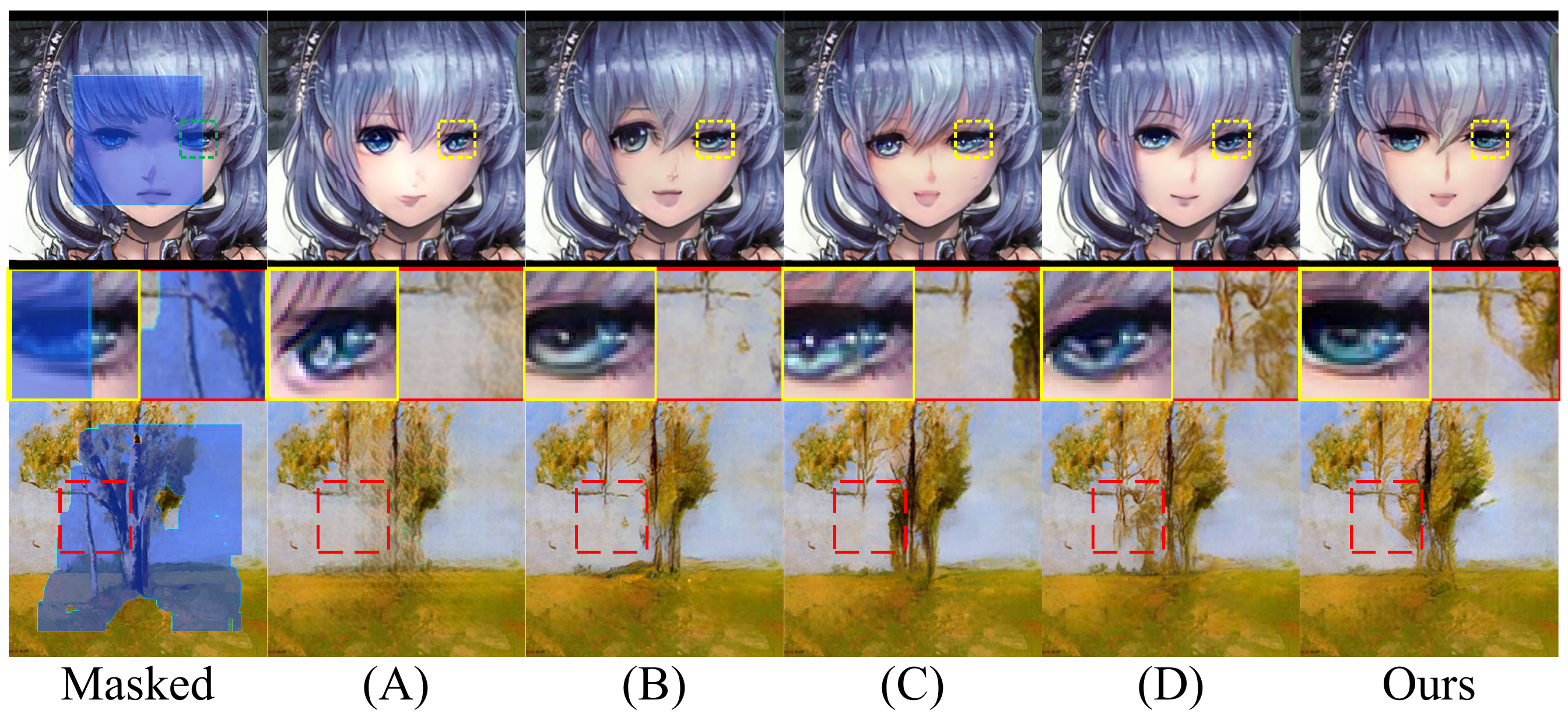}
	\caption{Visual examples of the ablation study with  (A) GRIG without the forgery-patch discriminator, (B) GRIG without the projected discriminator,  (C) GRIG replaced forgery-patch discriminator with PatchGAN's discriminator, (D) GRIG replaced forgery-patch discriminator with SM-PatchGAN's discriminator, and GRIG (ours).}\label{fig:fig_components_ablations}
\end{figure}

Fig.~\ref{fig:fig_large_free_form_img} shows the corresponding qualitative performance. It demonstrates that semantic inpainting on large masks remains difficult for most inpainting methods. RFR~\cite{Li2020} produces repetitive image patches in inpainted regions because iterative refinement in feature space may overlook fine details in image space. 
AOT~\cite{Zeng2021} and CMOD~\cite{Zhao2021} perform well on these datasets. However, with the complex background, they would struggle for larger masked areas in some cases. Because the one-time inference cannot re-adjust inpainted results, the complex background may likely have negative impacts on MAT's inpainting quality. Because fine details are easily overlooked in projected features, projected-based models~\cite{Sauer2021} tend to produce blurry results. 
Overall, the proposed method, Lama~\cite{Suvorov2022}, and MAT~\cite{Li2022} can inpaint plausible contents in complex structures with high mask ratios.

\begin{table}[t]
	
	\centering
    \setlength{\tabcolsep}{3.6mm}
    \caption{Quantitative results (FID) of the ablation study with (A) GRIG without the forgery-patch discriminator, (B) GRIG without the projected discriminator, (C) GRIG replaced forgery-patch discriminator with PatchGAN's discriminator, (D) GRIG replaced forgery-patch discriminator with SM-PatchGAN's discriminator, and GRIG (ours). Results were evaluated on  $50\!-\!60\%$ mask ratios. \textbf{Bold}: top-1 best quantity.}
 {
		\begin{tabular}{c|ccccc}
			\hline
			Dataset  &(A)  &(B)&(C)&(D)&Ours \\
            \hline
            CHASE & 73.96& 57.00 & 56.89  & 64.17   & \textbf{55.84}\\
            \hline
			Anime&  77.49 & 68.56 & 66.03 &  71.12 & \textbf{65.05}\\
            \hline
            Dog&   65.33 & 62.92&  61.17    &  59.83  &   \textbf{58.49}\\
            \hline
            Art&  96.84 & 79.19  & 79.16 &   78.83 &  \textbf{77.32}\\
            \hline
			CelebA-HQ   & 10.14   &  8.92 &  8.41  & 8.96   & \textbf{8.06}   \\
            \hline
			PSV   &  60.53     & 61.76 & 59.62 & 61.07    & \textbf{58.08}     \\
			
			\hline
	\end{tabular}}
	\label{tab:components_ablations}
\end{table}

\begin{figure*}[!thbp]
	\centering
	\includegraphics[width=0.92\textwidth]{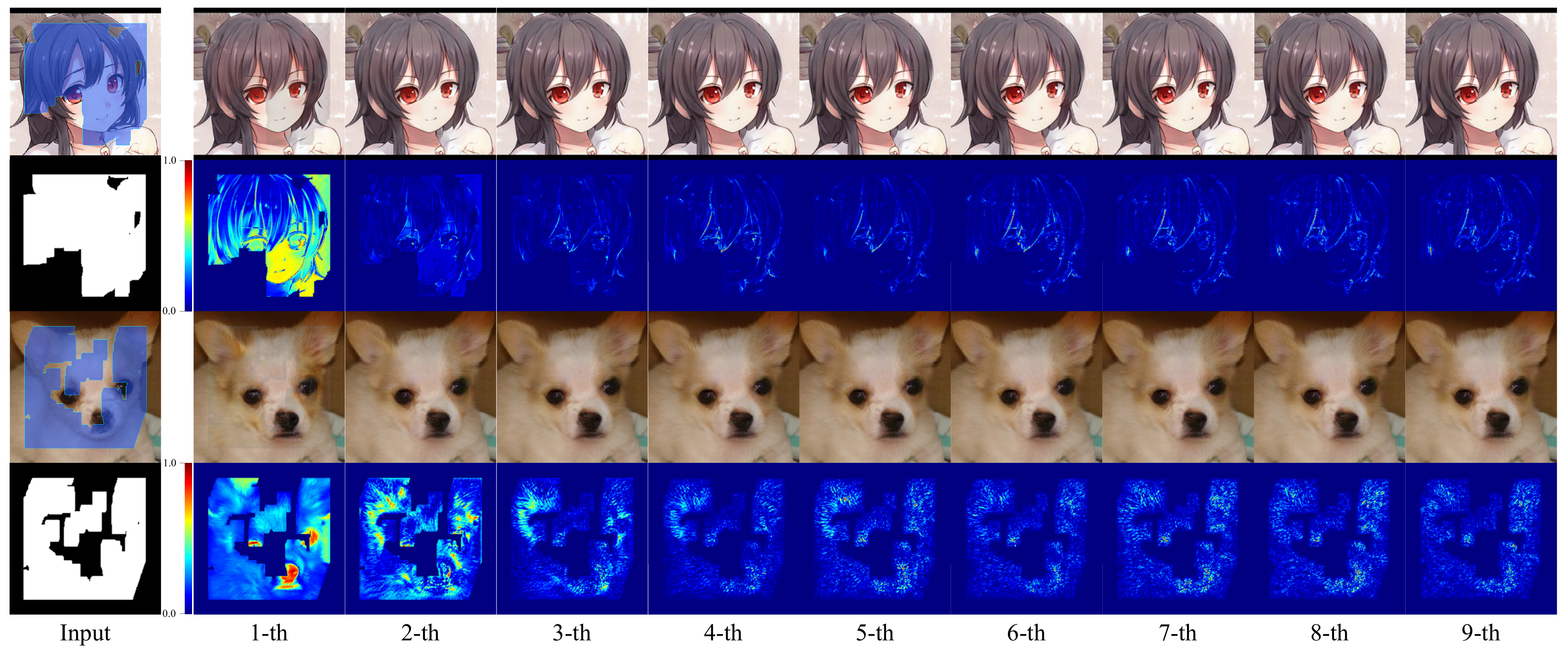}
	\caption{Visualization of our inpainting performance at each iterative reasoning step \revise{(for each group)}: (top-left) masked image, (bottom-left) input binary mask, (top-right) inpainted images, and (bottom-right) heatmaps of residual outputs {$\Delta_t$}. Color in red/blue in heatmaps represents the high/low change for the corresponding pixel.}\label{fig:fig_n_iter_show_imgs}
\end{figure*}

\subsection{Ablation study}

\subsubsection{Ablation study on discriminators}
We further analyzed the \revise{effects of discriminators in GRIG by individually removing each component and replacing our forgery-patch discriminator with PatchGAN~\cite{Isola_2017_cvpr} and SM-PatchGAN~\cite{Zeng2021}, respectively. All compared discriminators used the same network architecture of $70\times 70$-sized PatchGAN.  We evaluated their inpainting performance to show the impact of these changes.}

\revise{Table~\ref{tab:components_ablations} shows the quantitative results of compared variants.  GRIG outperforms all variants in terms of the FID score on differing few-shot and large-scale datasets. The FID scores increase dramatically when either the projected discriminator or the forgery-patch discriminator is removed. Replacing our forgery-patch discriminator with other discriminators also leads to higher FID scores. These results  indicate that removing our discriminators or replacing the proposed forgey-patch discriminator causes a rapid degradation in inpainting performance. The best FID scores of our GRIG on various datasets validate the effectiveness of our forgery-patch discriminator for performance boosting and  mitigating overfitting on few-shot image inpainting.
 }

Fig.~\ref{fig:fig_components_ablations} shows corresponding visual comparisons. 
When removing the forgery-patch discriminator, the inpainted results show noticeable artifacts around mask boundaries, and these produced textures are blurry, as shown in Fig.~\ref{fig:fig_components_ablations}A.  When the projected discriminator is removed, both quantitative performance and visual quality suffer. It would be more difficult to maintain the semantic structure of outputs in this case, e.g., the asymmetrical anime face, as shown in Fig.~\ref{fig:fig_components_ablations}B. The alignment between generated pixels and known pixels may be influenced when we replace our \revise{forgery-patch} discriminator with a Patch-GAN discriminator, as shown in Fig.~\ref{fig:fig_components_ablations}C. When we replace our \revise{forgery-patch} discriminator with an SM-PatchGAN discriminator, it can create plausible contents, but the consistency with known areas is poor, as seen in Fig.~\ref{fig:fig_components_ablations}D. In comparison, GRIG shows the best performance on both quantitative and qualitative measures.



\begin{table}[t]
	\centering
    \setlength{\tabcolsep}{3.3mm}
    \caption{Quantitative results (FID) of the models trained with various iterative reasoning steps $T$. Results were evaluated on  $50\!-\!60\%$ mask ratios. \textbf{Bold}: top-1 best quantity.}
 {
		\begin{tabular}{c|ccccc}
			\hline
			Dataset &$T=1$  &$T=3$  &$T=5$&$T=7$&$T=9$\\
            \hline
			CHASE   &   62.76    &    55.84     &   56.58     &    \textbf{53.39}      &      57.84    \\
            \hline
			Anime    &     69.55   &  \textbf{65.05}     & 68.05     &  66.50   &   69.20    \\
	        \hline		
            Dog   &    63.64  &  \textbf{58.49}   &   59.66   &   62.09    &     61.47    \\
	        \hline		
            Art   &    78.40  &   \textbf{77.32}     &  78.04   &   78.29     &   78.23      \\
			\hline
	\end{tabular}}
	\label{tab:iter_model_performance}
\end{table}

\begin{figure}[t]
	\centering
	\includegraphics[width=0.485\textwidth]{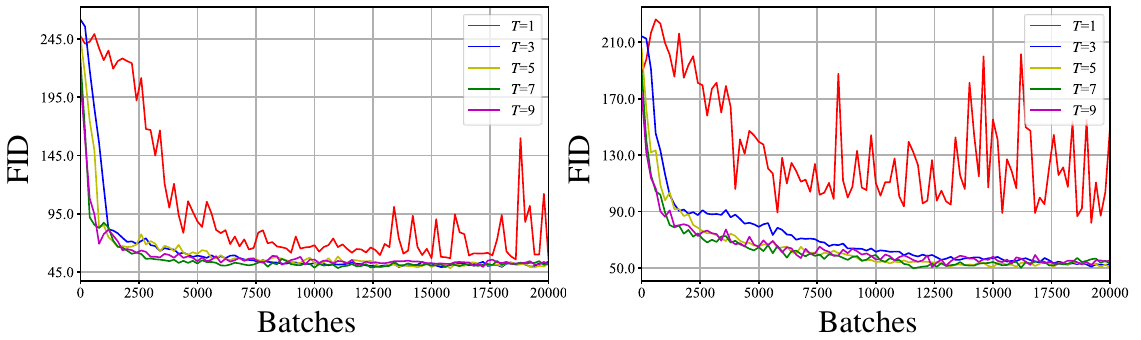}
	\caption{Comparisons of FID scores for each training iteration on  Anime (left) and Dog (right) datasets. Results were evaluated on the fixed center 25\% rectangular mask.}\label{fig:fig_train_iter_fid}
\end{figure}

\subsubsection{Ablation study on iterative reasoning steps}
To evaluate the effectiveness of the iterative reasoning steps $T$, we \wanglong{performed an ablation study for GRIG with various iterative reasoning steps $T$} and tested FID scores on $50\!-\!60\%$ mask ratios, respectively, as shown in Table~\ref{tab:iter_model_performance}. \wanglong{Each test used the same iterative reasoning steps to the corresponding training phase.}  Compared to models trained on $T=1$, models trained on $T>1$ have large performance gains. For example,  on CHASE dataset, the model trained on $T=3$ is 6.92 lower of the FID score than that trained on $T=1$ (55.84 vs 62.76).  When $T>5$, the performance gains are saturated or decreased to some extent, but the inpainting performance is still better than $T=1$ in most cases. \wanglong{The results indicate that GRIG can produce satisfactory inpainting outcomes in the early steps, while the residual offsets may fluctuate in subsequent steps, potentially leading to variations in inpainting quality. However, GRIG effectively balances the number of steps and the improvement in inpainting quality, achieving superior performance in the few-shot image inpainting task.} In this paper, we used $T=3$ to strike a balance between computational cost and visual quality. 
Fig.~\ref{fig:fig_n_iter_show_imgs} \revise{visualizes} the residual output $\Delta_t$ for each step $t$ for the model trained with $T=3$. The masked images were gradually inpainted. The model prioritizes semantic features in the early steps and fine details in the later steps.





{As shown in Fig.~\ref{fig:fig_train_iter_fid}, we evaluated our GRIG on the fixed center 25\% rectangular mask for models trained with $T=1,3,5,7,9$, respectively.} The FID scores on Anime and Dog datasets show that models trained with higher iterative reasoning steps $T$ converge faster than those with lower $T$, and models trained with $T=1$ cannot converge easily. Specifically, models trained with  $T>1$ converge around $10,000$ image batches, whereas models trained with $T=1$ are far from convergence and fluctuate drastically even after $10,000$ image batches. {It shows that our framework can effectively help networks to converge faster.}

\subsubsection{Ablation study on few-shot settings}
\wanglong{We tested the sensitivity of GRIG to different few-shot settings on CHASE, Anime, Cat, Dog, and Art datasets.} The quantitative results of FID scores are shown in Table~\ref{tab:n_shot_ablations}. FID scores decrease as the number of training samples increases (e.g., $50$-shot images), implying that more training samples could improve inpainting quality.

\begin{figure*}[t]
	\centering
	\includegraphics[width=\textwidth]{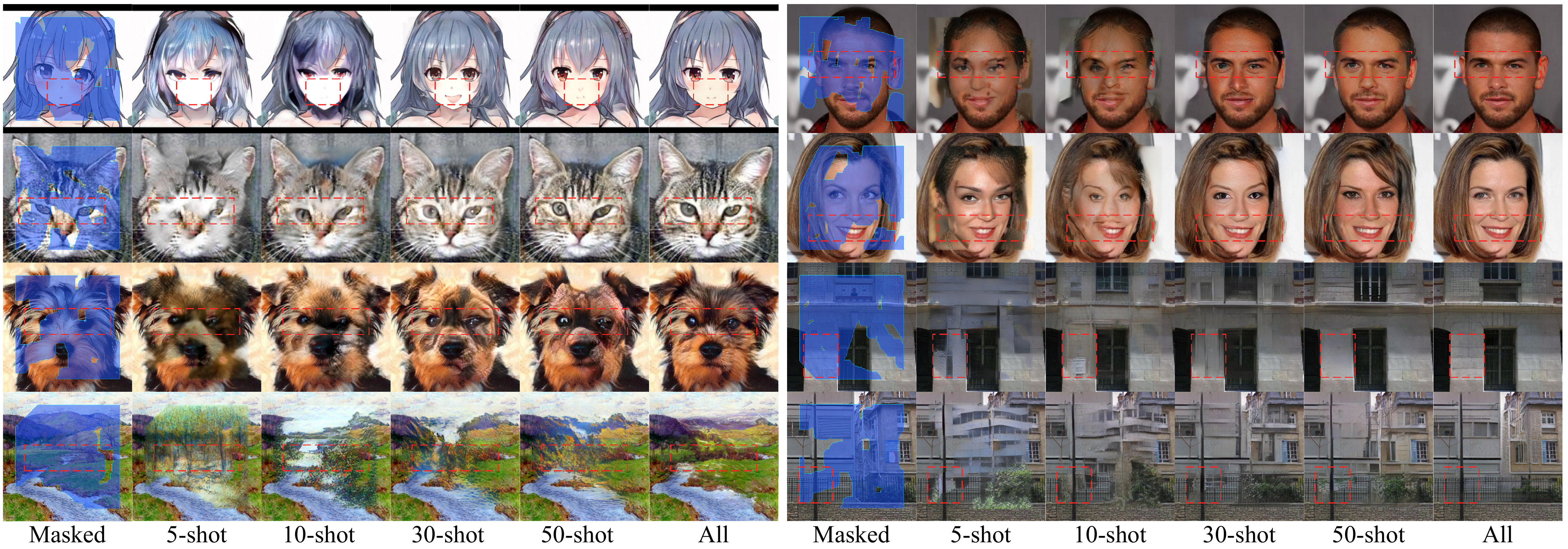}
	\caption{Visual comparisons between models trained on different $n$-shot settings. ``All'' means the training sets mentioned in the paper.   }\label{fig:fig_n_shot_ablation_imgs_all}
\end{figure*}

Fig.~\ref{fig:fig_n_shot_ablation_imgs_all} presents the corresponding inpainted examples. The quality of inpainted images drops quickly when models were trained on fewer samples. For example, models trained on $5$-shot images are unable to inpaint semantic structures within masked areas; while models trained on $10$-shot and $30$-shot  images can inpaint more plausible contents, some output results still show obvious color inconsistency around mask boundaries. A similar phenomenon is also shown on CMOD and MAT in Fig.~\ref{fig:fig_inpainting_performance_sub_b}.  In contrast, models trained on $50$-shot settings  produce sharper results with more complex textures and rich colors. We can find that the more training samples the models trained on, the better their performance on both quantitative and qualitative evaluations.

\section{Conclusion, limitations, and future work}
\label{conclusion} 
We took a first step toward solving few-shot image inpainting in this paper. \wanglong{By introducing iterative residual reasoning} \revise{with decoupled image-level and patch-level discriminators,} \wanglong{we presented a novel few-shot generative residual image inpainting framework. The proposed generator effectively utilizes CNNs for feature extraction and Transformers for global reasoning.} To assist the generative network in learning image fine details, a forgery-patch discriminator was introduced. Furthermore, we established the new state-of-the-art performance on multiple few-shot datasets, and extensive experiments demonstrated the efficacy of the proposed method.

\begin{table}[t]
	\centering
    \setlength{\tabcolsep}{2.9mm}
    \caption{Quantitative results (FID) of the models trained on different $n$-shot settings. Results were evaluated on  $50\!-\!60\%$ mask ratios. The term ``$n$-shot'' means that we used $n$ images for training while ``All'' means the training sets mentioned in this paper. \textbf{Bold}: top-1 best quantity.}
 {
		\begin{tabular}{c|ccccc}
			\hline
			Dataset &$5$-shot  &$10$-shot  &$30$-shot&$50$-shot&All\\
            \hline
			CHASE   &  70.26   &    64.39    & -    & -    & \textbf{55.84}      \\
	        \hline		
            Anime   &   98.50   &   90.02   & 77.58   &  77.32    & \textbf{65.05}  \\
	        \hline		
            Cat   &    120.42  &  73.10  &   58.93  &  55.38   &    \textbf{44.73}     \\
			\hline
            Dog   &    138.12    &126.37    &  98.83  & 88.63   &  \textbf{58.49}   \\
	        \hline		
            Art   &  113.27   &  102.02   & 96.17   &  91.31   &  \textbf{77.32}   \\
            \hline	
            CelebA-HQ     &  52.46    &  31.78  & 20.06    &   15.53      &    \textbf{8.06}       \\
            \hline
			PSV     &    122.23 &  105.09  &  97.33 &    90.23  &  \textbf{58.08}       \\
			\hline
	\end{tabular}}
	\label{tab:n_shot_ablations}
\end{table}

Our method has some limitations. The approach can effectively perform high-fidelity image inpainting on few-shot datasets. However, GRIG cannot directly utilize conditional information for guidance-based image inpainting. Introducing a more sophisticated scheme or module to guide the inpainting process would be more interesting for controllable few-shot image completion. Moreover, GRIG is not specialized in diverse image inpainting. Using a mapping network to embed random style codes into the generator could be a good solution for the diversity of few-shot image inpainting.


{
\bibliographystyle{IEEEtran}
\bibliography{IEEEabrv,inpainting_ijcai_bib}
}

\end{document}